\documentclass[11pt,a4paper]{article}
\usepackage{times,latexsym}
\usepackage{url}
\usepackage[T1]{fontenc}
\usepackage{amsmath}
\usepackage{enumitem}
\usepackage{multirow}
\usepackage{comment}
\usepackage{capt-of}
\usepackage{microtype}
\usepackage{graphicx}
\usepackage{subfigure}
\usepackage{booktabs} %
\usepackage{multirow}
\usepackage{xcolor,colortbl}
\usepackage{amsfonts}
\usepackage{arydshln}
\usepackage{xspace}

\usepackage[acceptedWithA]{tacl2021v1}

\usepackage{xspace,mfirstuc,tabulary}

\newif\iftaclinstructions
\taclinstructionsfalse %
\iftaclinstructions
\renewcommand{\confidential}{}
\renewcommand{\anonsubtext}{(No author info supplied here, for consistency with
TACL-submission anonymization requirements)}
\newcommand{\instr}
\fi

\iftaclpubformat %

\else

\fi

\newcommand{\E}{{\mathbb{E}}}

\newcommand{\LL}{{\mathcal{L}}}
\newcommand{\LDAT}{{\mathcal{L}}_{\text{DAT}}}

\newcommand{\bV}{{V}}
\newcommand{\bv}{\mathbf{v}}

\newcommand{\R}{\mathbb{R}}

\newcommand{\texterr}[1]{\textit{\textcolor{red}{#1}}}

\newcommand{\methodname}{PreDAT\xspace}

\definecolor{mypurple}{HTML}{7030A0}
\definecolor{mygreen}{HTML}{385723}
\definecolor{myorange}{HTML}{ED7D31}
\definecolor{myblue}{HTML}{2F5597}
\definecolor{mygray}{HTML}{7F7F7F}
\definecolor{LightCyan}{rgb}{0.88,1,1}

\newcommand{\textgood}[1]{{\textcolor{blue}{#1}}}
\newcommand{\textunrel}[1]{{\textcolor{brown}{\textit{#1}}}}

\newcommand{\omitone}[1]{}
\newcommand{\omittwo}[1]{}

\DeclareUrlCommand\Code{\urlstyle{rm}}
\expandafter\def\expandafter\UrlBreaks\expandafter{\UrlBreaks  
\do\/\do\a\do\b\do\c\do\d\do\e\do\f\do\g\do\h\do\i\do\j\do\k
\do\l\do\m\do\n\do\o\do\p\do\q\do\r\do\s\do\t\do\u\do\v
\do\w\do\x\do\y\do\z
\do\A\do\B\do\C\do\D\do\E\do\F\do\G\do\H\do\I\do\J\do\K
\do\L\do\M\do\N\do\O\do\P\do\Q\do\R\do\S\do\T\do\U\do\V
\do\W\do\X\do\Y\do\Z}

\title{Directed Acyclic Transformer Pre-training for High-quality Non-autoregressive Text Generation}

\author{
  \textbf{Fei Huang}
  \quad
  \textbf{Pei Ke}
  \quad
  \textbf{Minlie Huang\Thanks{Corresponding author: Minlie Huang.}}
  \\
  The CoAI group, Tsinghua University, Beijing, China \\
Institute for Artificial Intelligence, State Key Lab of Intelligent Technology and Systems, \\
Beijing National Research Center for Information Science and Technology, \\
Department of Computer Science and Technology, Tsinghua University, Beijing, China \\
  \small{\texttt{f-huang18@mails.tsinghua.edu.cn}, \texttt{kepei1106@outlook.com}} \\
  \small{\texttt{aihuang@tsinghua.edu.cn}}
}

\date{}

\begin{document}
\maketitle
\begin{abstract}
Non-AutoRegressive (NAR) text generation models have drawn much attention because of their significantly faster decoding speed and good generation quality in machine translation.
However, in a wider range of text generation tasks, existing NAR models lack proper pre-training, making them still far behind the pre-trained autoregressive models.
In this paper, we propose Pre-trained Direct\-ed Acyclic Transformer (\methodname) and a novel pre-training task
to promote prediction consistency in NAR generation.
Experiments on five text generation tasks show that our \methodname remarkably outperforms existing pre-trained NAR models (+4.2 scores on average) and even achieves better results than pre-trained autoregressive baselines in n-gram-based metrics, along with 17 times speedup in throughput.
Further analysis shows that \methodname benefits from the unbiased prediction order that alleviates the error accumulation problem in autoregressive generation, which provides new insights into the advantages of NAR generation.\footnote{Our codes and pre-trained models are available at \url{https://github.com/thu-coai/DA-Transformer}.}
\end{abstract}

\section{Introduction}
\label{sec:intro}
\addtolength{\skip\footins}{-0.2em}
\addtolength{\abovedisplayskip}{-0.15em}
\addtolength{\belowdisplayskip}{-0.15em}

Pre-trained language models have been widely applied in text generation~\cite{gpt2_2019, mass2019song, bart2020lewis, t5_2020}, which can effectively improve the performance of downstream generation tasks, especially in low-resource scenarios~\cite{gpt3_2020}.
Most of these pre-trained language models are based on AutoRegressive (AR) generation, which produces high-quality texts by predicting each token one by one.
However, such a sequential generation process suffers from high latency and low throughput in inference, thereby largely limiting the use of AR models in scenarios with real-time requirements.

Non-AutoRegressive (NAR) generation is an al\-ter\-native text generation paradigm~\cite{nat2018gu}.
Unlike sequential generation in AR models, NAR models predict all tokens in parallel, which largely accelerates the decoding process.
Although early NAR models suffer from serious quality degradation due to the independent token prediction, recent NAR studies have made much progress on some generation tasks, such as machine translation~\cite{glat2021qian, tricktrade2021gu, dslp2021huang}.
Notably, \citet{dat2022huang} propose Directed Acyclic Transformer, which incorporates a directed acyclic graph to reduce the conflicts in capturing possible outputs, 
achieving a comparable translation quality to the AR models.

Despite the success of NAR generation in machine translation, it is still challenging to apply NAR models to a wider range of generation tasks,
mainly due to the lack of appropriate pre-training. \\
Although some previous studies have explored pre-training methods such as directly fine-tuning BERT for NAR generation~\cite{natbert2020, nagcrf2021su, mist2021jiang} or pre-training NAR models from scratch~\cite{bang2021qi, elmer2022li},
their models still have a significant quality gap compared with AR ones.
We argue that these methods do not fully exploit the characteristic of NAR generation, thereby restricting downstream performance.
Specifically, we discuss two main issues:
(1) Previous pre-training tasks are ineffective in promoting sentence-level prediction consistency, making it hard for their models to predict a whole sentence simultaneously while preserving the fluency in downstream NAR generation.
(2) Previous pre-training tasks fail to address the multi-modality problem~\cite{nat2018gu}, which has proved to be a fundamental and important challenge in training NAR models~\cite{mple2021}.

In this paper, we introduce \methodname, a Pre-trained Directed Acyclic Transformer for high-quality non-autoregressive text generation.
We utilize the architecture of Directed Acyclic Transformer
and further propose a novel pre-training task, Double-Source Text Infilling (DSTI), aiming to address the above issues in pre-trained NAR models.
Specifically, DSTI contains two steps:
it corrupts a sentence and scatters the tokens into two sequences, which are fed into the encoder and decoder as two sources of information;
then the model is trained to recover the corrupted fragments non-autoregressively.
During the pre-training, our model predicts long sentence fragments (about 15 tokens) from nearby contexts, which promotes prediction consistency and bidirectional dependencies.
Moreover, DSTI designs a strategy for creating pre-training data pairs that allow the output sequences to have flexible lengths, which well incorporates various alignment-based NAR training objectives to alleviate the multi-modality problem~\cite{ctc2018libovicky,axe2020ghazvininejad,oaxe2021du,dat2022huang}.

Automatic evaluation shows that \methodname is effective and efficient on five text generation tasks. It remarkably outperforms previous pre-trained NAR models (+4.2 scores on average) and even achieves better results than pre-trained AR baselines in n-gram-based metrics (+0.7 scores on average), along with a 17x speedup in throughput.
To our knowledge, \methodname is the first NAR model that outperforms pre-trained AR models on various generation tasks in automatic evaluation.
Further ablation studies verify that our pre-training task designs, including the long fragment prediction and alignment-based training objectives, are crucial for success.

To better understand the advantages and weaknesses of NAR generation, we use automatic and manual methods to investigate the generated texts in downstream tasks.
We find that \methodname can alleviate the error accumulation in AR generation and improve the relevance to the input, thereby leading to a better performance in n-gram-based metrics.
However, we also find that NAR models, including \methodname, are still weaker than AR models in preserving the consistency among generated tokens, leading to grammatical errors such as wrong word choices.
We believe that these findings can provide novel insights for future NAR studies.

\section{Related Work}

\noindent\textbf{Pre-trained Language Models (PLM)}\quad
In recent years, PLMs have made significant progress in natural language generation~\cite{gpt2_2019, mass2019song, bart2020lewis, t5_2020}.
These PLMs are pre-trained on a large corpus of unlabeled data, where the knowledge can be transferred to downstream tasks, resulting in improved generation quality.

\vspace{0.2em}
\noindent\textbf{Non-Autoregressive Generation}\ \ 
Although NAR generation~\cite{nat2018gu} remarkably speeds up the inference, \citet{mple2021} point out that it theoretically suffers from serious information dropping, previously known as the multi-modality problem. To alleviate the problem, previous studies propose methods including (1) iterative refinement~\cite{iterativerefinement2018lee, levenshtein2019gu, cmlm2019ghazvininejad, jmnat2020guo, cmlmc2021}; (2) knowledge distillation~\cite{seqkd2016kim, monokd2022ding, parakd2021ding, lexical2021ding, diversekd2022shao}; (3) dependency enhancement~\cite{crf2019sun, glat2021qian, dslp2021huang, latentglat2022bao}; (4) alignment-based objectives~\cite{axe2020ghazvininejad, oaxe2021du, ctc2018libovicky, dat2022huang}.

There are also studies combining PLMs and NAR generation.
For example, some methods
fine-tune existing pre-trained models directly~\cite{mist2021jiang} or with an adapter~\cite{natbert2020, nagcrf2021su}.
Some others combine AR and NAR prediction~\cite{bang2021qi} or involve an early exiting mechanism~\cite{elmer2022li} in pre-training.

Compared with these studies, our method has two significant differences:
(1) Previous methods either predict short spans (e.g., BERT) or incorporate unidirectional AR prediction (\citealp{bang2021qi}), which hardly contribute to NAR generation that predicts a whole sentence with bidirectional attention.
In contrast, we train our model to predict long fragments simultaneously, leading to better consistency among generated tokens.
(2) Previous methods use a token-level loss 
that forces the model to predict a same-length sequence to match the target,
which over-penalizes the position shift error \cite{axe2020ghazvininejad} and worsens the multi-modality problem.
We introduce an up-sampling strategy to obtain longer output sequences, which well incorporates previous alignment-based NAR losses to address the above problems.

\section{Preliminaries: Directed Acyclic Transformer}
\label{sec:preliminary}

\begin{figure}[t]
\centering
\includegraphics[width=0.9\linewidth]{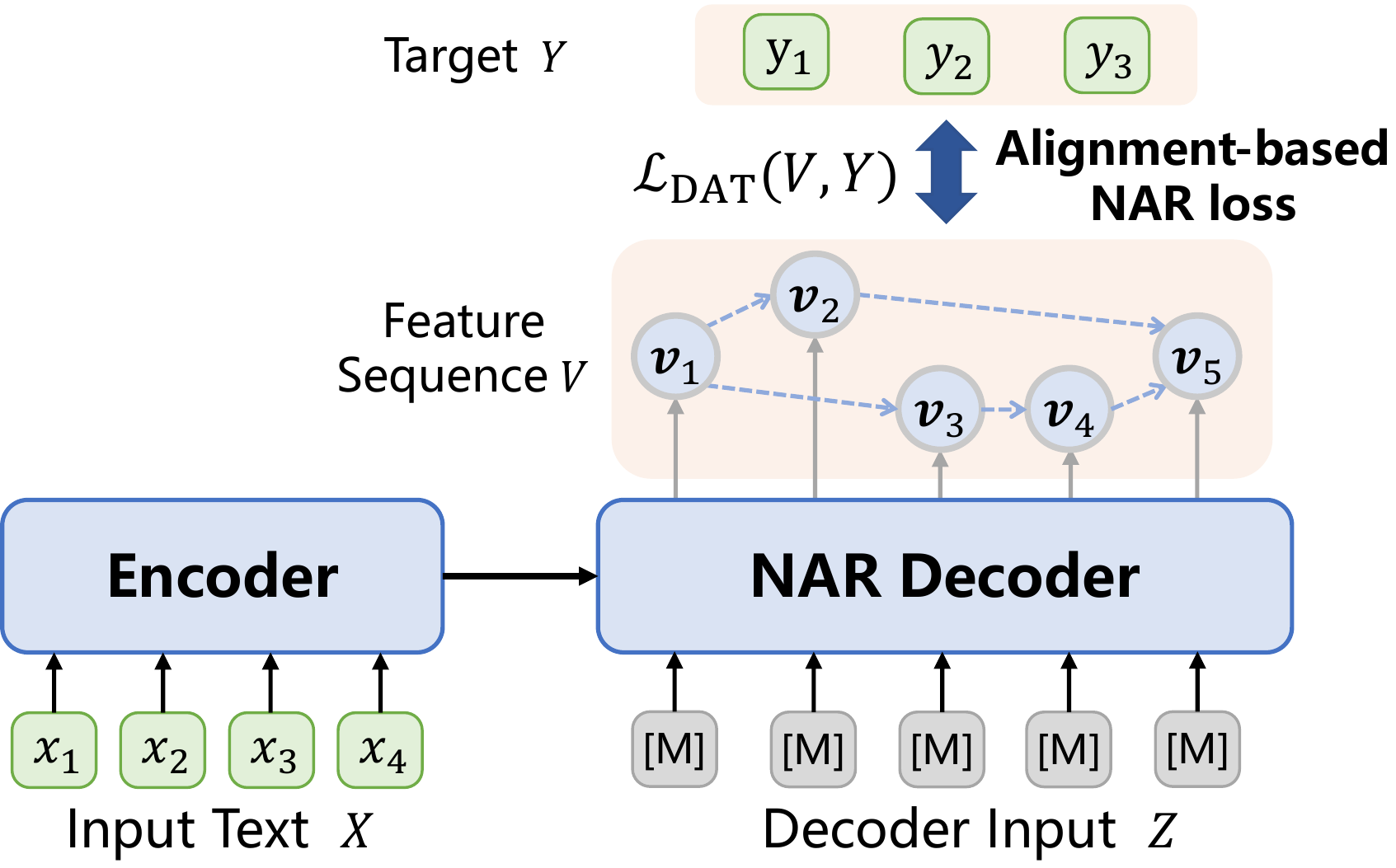}
\vspace{-0.7em}
\caption{Preliminaries: Directed Acyclic Transformer (DAT). To alleviate the multi-modality problem, DAT predicts a feature sequence $\bV$ organized in a directed acyclic graph (DAG) and then adopts an alignment-based objective that aligns the target $Y$ to the feature sequence $\bV$, represented by $\LDAT(\bV, Y)$.}
\vspace{-0.8em}
\label{fig:dat}
\end{figure}

\begin{figure*}[!t]
\centering
\vspace{-0em}
\includegraphics[width=0.94\linewidth]{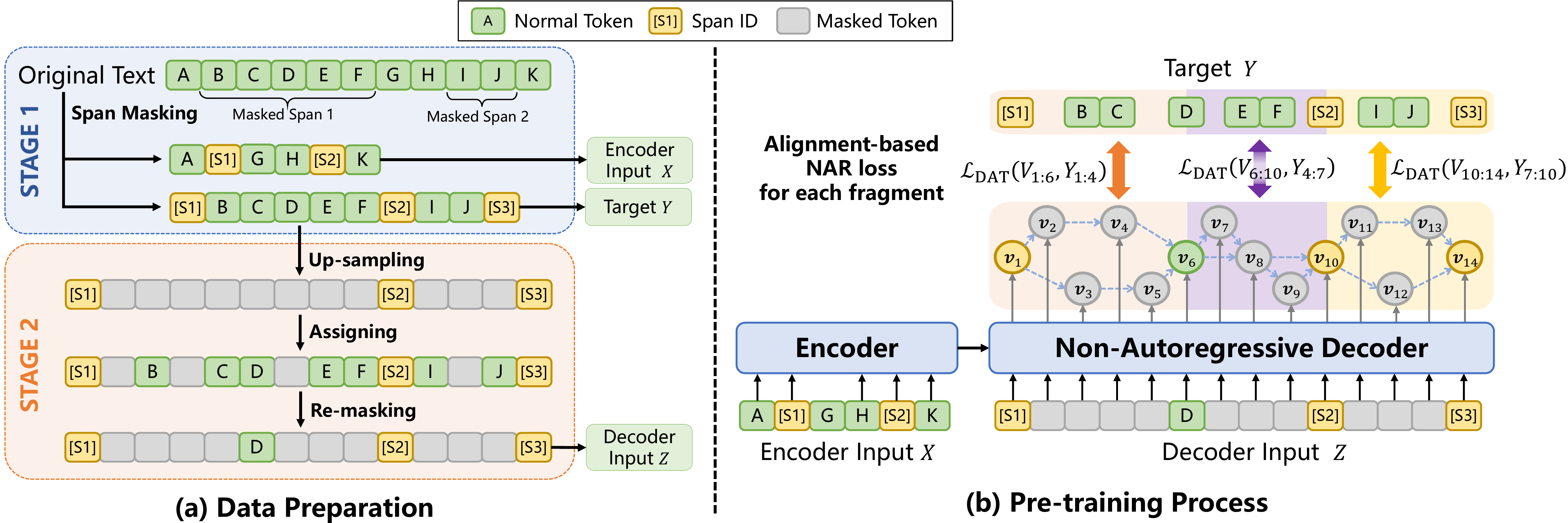}
\vspace{-0.5em}
\caption{An overview of Double-Source Text Infilling (DSTI). (a) Data Preparation: DSTI first creates the encoder input $X$ and the target $Y$ by span masking, and then obtains the decoder input $Z$ by up-sampling, assigning, and re-masking.
(b) Pre-training Process: The NAR model is trained to predict the unseen fragments in $Y$ in parallel, with $X$ and $Z$ as inputs. The training objective is the sum of alignment-based NAR losses, which are obtained by aligning each target fragment (e.g., $Y_{1:4}$) to the feature sequence on the corresponding masked segments (e.g., $\bV_{1:6}$).}
\vspace{-1em}
\label{fig:overview}
\end{figure*}

Directed Acyclic Transformer (DAT, \citealp{dat2022huang}) is an NAR model that effectively alleviates the multi-modality problem. It introduces a longer decoding sequence and an alignment-based objective to reduce the conflicts in capturing multiple possible outputs.
Specifically, given the input $X = \{ x_1, \cdots, x_M \}$ and the target sequence $Y = \{ y_1, \cdots, y_N \}$, DAT produces a feature sequence $\bV = \{ \bv_1, \bv_2, \cdots, \bv_L \}$ organized in a Directed Acyclic Graph (DAG), where $Y$ is aligned to a sub-sequence of $V$ (equivalently, assigned to a path of the DAG). Notably, $L$ is usually much larger than $N$ to allow for more flexible alignments.
In DAT training, the alignment-based objective marginalizes the probabilities of all possible alignments that produce the target $Y$, formally as

\vspace{-1em}
{
\small

\begin{align}
    \LDAT(\bV, Y) &= -\log P_{\theta}(Y|X) \notag \\
    &= -\log \sum_{A \in \Gamma} P_{\theta}(Y|A, X) P_{\theta}(A|X), \label{eq:LDAT} \\
    P_{\theta}(Y|A, X) &= \prod\nolimits_{i=1} ^ {L} P_{\theta}(y_i | \bv_{a_{i}}), \notag \\
    P_{\theta}(A|X) &= \prod\nolimits_{i=1} ^ {L-1} P_{\theta}(a_{i+1} | a_{i}), \notag
\end{align}

}
\vspace{-0.0em}

\noindent where $A = \{ a_1, \cdots, a_{L}\}$ is feature indexes on the aligned path, and $\Gamma$ contains all possible paths with the size of $\tbinom{L}{N}$. $P_{\theta}(y_i | \bv_{a_{i}})$ represents {token} probabilities predicted from the feature $\bv_{a_{i}}$, and $P_{\theta}(a_{i+1} | a_{i})$ represents transition probabilities revealing how likely $a_{i+1}$ follows $a_i$ in a path.
Since it is impossible to enumerate the huge number of paths in Equation \ref{eq:LDAT}, a dynamic programming algorithm can be adopted to address the problem, whose details can be found in the original paper \cite{dat2022huang}.

Compared with previous NAR models, DAT explicitly models the dependencies between tokens by the position transitions and is able to store multiple modalities on different paths of the DAG, thereby remarkably improving the generation performance. Moreover, various decoding algorithms such as beam search and Nucleus sampling \cite{topp2020holtzman} can be utilized to boost the generation quality or diversity.

Besides $\LDAT$, there are other alignment-based objectives that succeed in alleviating the multi-modality problem in NAR generation, such as AXE \cite{axe2020ghazvininejad}, OaXE \cite{oaxe2021du}, and CTC \cite{ctc2006graves,ctc2018libovicky}. In general, these objectives are also obtained by aligning the target $Y$ with the feature sequence $\bV$, thus denoted by $\LL(\bV, Y)$.

\section{Proposed Method}

In this section, we introduce \methodname, Pretrained Directed Acyclic Transformer.
We first propose the pre-training task (Sec.\ref{sec:pretrain}) and then describe the fine-tuning and inference strategies (Sec.\ref{sec:finetune}).

\subsection{Pre-training Task}
\label{sec:pretrain}

Our pre-training task, Double-Source Text Infilling (DSTI), is a self-super\-vised pre-training task that aims to promote prediction consistency and bidirectional dependencies for NAR models.
Our task scatters part of a sentence into two sequences, feeds them into the encoder and decoder as two sources of information, and then trains the model to predict long unseen fragments in a non-autoregressive fashion.
Although DSTI is compatible with various NAR architectures and losses, we mainly focus on DAT due to its superior performance.

As shown in Fig.\ref{fig:overview}, our task takes a piece of text from the pre-training corpus and decomposes it into a triple $(X, Z, Y)$, where $X = \{ x_1, \cdots, x_M \}$ is the encoder input, $Z = \{ z_1, \cdots, z_L \}$ is the decoder input, and $Y = \{ y_1, \cdots, y_N \}$ is the target. The data preparation consists of two stages.

\vspace{0.2em}
\noindent\textbf{Stage 1: Creating Encoder Input}\quad
We utilize span masking~\cite{t5_2020} to obtain the encoder input $X$ and the target $Y$. 
Specifically, we randomly mask tokens in the original sentence, and then replace consecutive masks into a single special token representing the span ID.
Then the prediction target $Y$ is constructed by concatenating the masked spans with the span IDs as delimiters. %

Specially, we force each masked span to be long enough (about 15 tokens) because the NAR model has to generate a whole sentence simultaneously in inference, where predicting short spans is unhelpful in preserving sentence-level consistency.

\vspace{0.2em}
\noindent\textbf{Stage 2: Creating Decoder Input}\quad
The decoder input $Z$ plays two roles in our pre-training: (1) It reveals some target tokens to promote bidirectional dependencies in the decoder.
(2) It determines the length of the predicted feature sequence.

To incorporate the alignment-based NAR losses that require a longer feature sequence than the target (such as DAT and CTC), we create the decoder input $Z$ by an up-sampling step. 
Then we assign a part of the target tokens to appropriate positions in $Z$, where the unseen tokens will be used as prediction targets.
Specifically, creating $Z$ follows three steps: up-sampling, assigning, and re-masking.

For \textbf{up-sampling}, we decide the length of $Z$ based on the target length. Formally, we have $L := \lambda N$, where $\lambda$ is an up-sampling ratio.
In DAT, varying $L$ can bring different DAG sizes and structures, where we sample $\lambda$ from a uniform distribution
to diversify the DAG structures in pre-training.
After determining the length, the span IDs are put into $Z$ according to the up-sampling ratio, which will not be modified in the later steps.

For \textbf{assigning}, we distribute the target tokens in $Z$, regardless of whether the token will appear in the final input.
Formally, we use an assignment sequence $\{a_i\}_{1 \leq i \leq N}$ indicating that $z_{a_i} := y_i$. All other positions in $Z$ are masked.
For obtaining the sequence $\{a_i\}$, a straightforward strategy is to use uniform assignment, such that every two consecutive target tokens are separated by a constant number of [Mask].
In the pilot experiment, we find it better to use the strategy of glancing training~\cite{dat2022huang, glat2021qian}, which first predicts a DAG with a fully masked $Z$ and then assigns the target tokens on the positions that form the most probable path of the DAG.

For \textbf{re-masking}, we determine the tokens finally appeared in $Z$ and then mask the remaining ones.
Formally, we randomly sample a fixed proportion of tokens to form a set $R$, where $z_{a_i} := y_i$ if $i \in R$, and all the other tokens in $Z$ are masked.

\vspace{0.2em}
\noindent\textbf{Training Objective}\quad
Our objective is to reconstruct the unseen target fragments according to the given context, similar to masked language modelling \cite{bert2019devlin} but with a significant difference.
Instead of using a token-level loss that forces each masked position to predict the corresponding target token, we obtain the sum of alignment-based losses that aligns each unseen target fragment to the feature sequence predicted on the corresponding masked segments. Note that the feature sequence is longer than the target fragment, which brings a larger DAG with a higher capacity to capture multiple possible in-filling results.

Specifically, the decoder input consists of several consecutive masked segments segmented by the observed token or span IDs.
Each masked segment will produce a feature sequence $\bV_{a_i:a_j}$, which is then aligned to the corresponding target fragments $Y_{i:j}$ for the DAT loss.
The final loss is equal to the sum of the DAT loss of all fragments.
Formally,

\vspace{-1em}
{
\small
\begin{gather}
    \bV = \left[ \bv_1, \cdots, \bv_{|Z|} \right] = f_{\theta}(X, Z), \notag \\
    \LL  = \sum_{{i, j} \in \text{frag}(R)} \LL_\text{DAT}(\bV_{a_i:a_j}, Y_{i:j}), \notag
\end{gather}
}
%\vspace{-1em}

\noindent where $\text{frag}(R)$ consists of all pairs $(i, j)$ representing the start and end position of unseen fragments, and $\LDAT$ is defined in Equation \ref{eq:LDAT}.

Notably, our idea can be applied to other align\-ment-based NAR losses, such as CTC loss~\cite{ctc2006graves}, which also trains the model by aligning the target fragment to a longer predicted feature sequence. We verify the generality of DSTI with various loss functions in Sec.\ref{sec:ablation}.

\subsection{Fine-tuning and Inference}
\label{sec:finetune}

We generally follow the original training method \cite{dat2022huang} to fine-tune our \methodname on the downstream datasets while introducing some improvements:
we add a target length predictor for better adaption to tasks with various ratios of input and target lengths,
and further propose a trick to improve the decoding throughput.

\vspace{0.2em}
\noindent\textbf{Length Prediction}\quad
The original DAT simply sets the feature length $L$ to be a constant multiple of the input length, which in most cases of machine translation, satisfies the constraint that the feature sequence should be longer than the target length.
However, the targets in our downstream tasks can be arbitrarily long, making this strategy improper.

To better apply PreDAT to various generation tasks without the constraint of input and target length, we introduce a length predictor during fine-tuning and inference.
Specifically, in fine-tuning, we use a similar up-sampling strategy as the pre-training to obtain the decoder input length, i.e., $\lambda$ times the target length.
Then we adopt a length predictor on the top of the encoder and train it to predict the target length as a classification. In inference, we obtain the predicted length from the predictor, and then multiply it with $\hat{\lambda}$ to obtain the decoder input length, where $\hat{\lambda}$ is a hyper-parameter tuned on the validation set that controls the length of generated sentences.

\begin{figure}[t]
\centering
\vspace{-0em}
\includegraphics[width=\linewidth]{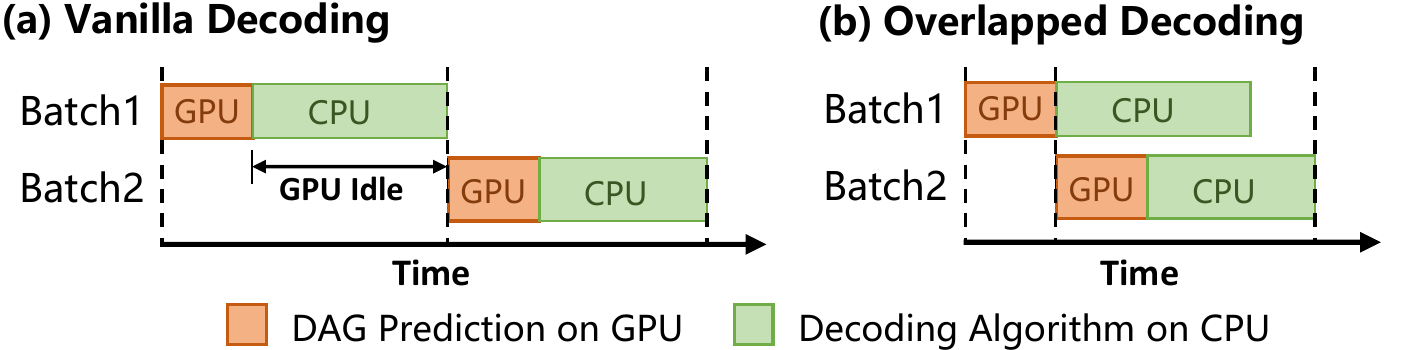}
\vspace{-1.5em}
\caption{Illustrations of (a) vanilla decoding and (b) overlapped decoding. Overlapped decoding reduces the GPU idle time, leading to higher decoding throughput.}
\vspace{-1em}
\label{fig:overlap}
\end{figure}

\vspace{0.2em}
\noindent\textbf{Overlapped Decoding}\ \ 
\methodname predicts the DAG in parallel on GPU, and then executes a decoding algorithm (e.g., beam search, \citealp{dat2022huang}) on CPUs to obtain the most likely output from the DAG.
As shown in Fig.\ref{fig:overlap}, we overlap the GPU and CPU execution, which reduces the GPU idle time and utilizes multiple CPU cores to parallelly process the batches, leading to remarkably higher decoding throughput while not affecting the latency.

\section{Experiments}

\subsection{Implementation Details}
\label{sec:implement}

\textbf{Model Configurations}\ \ 
Our \methodname is based on a 6-layer encoder-decoder Transformer~\cite{transformer2017vaswani} with a hidden size of 768, following the base version of AR and NAR baselines.

\vspace{0.2em}
\noindent\textbf{Pre-Training}\ \ 
We pretrain \methodname with Double-Source Text Infilling on 16GB English corpus from Wikipedia and BookCorpus~\cite{bookcorpus2015zhu}, with the vocabulary of \textit{bert-base-uncased}.
In stage 1, we take a sequence with about 600 tokens and mask 6 equal-length spans that account for 15\% tokens.
In stage 2, we sample $\lambda$ uniformly from $[4, 8]$ and mask 90\% tokens in the re-masking step.
Unless otherwise specified, we pre-train \methodname for 500k update steps with a batch size of 256 samples and use Adam optimizer \cite{adam2015kingma} with a learning rate of 2e-4.%
We utilize LightSeq~\cite{wang2021lightseq2} to accelerate the training (not used in inference), and the pre-training lasts approximately 72 hours on 8 Nvidia A100-40G GPUs.

\vspace{0.2em}
\noindent\textbf{Fine-Tuning}\ \ 
We fine-tune \methodname on downstream datasets with the DAT loss and glancing training \cite{glat2021qian, dat2022huang} without knowledge distillation.
According to the average sample lengths of each dataset, each mini-batch has approximately 4k target tokens for PersonaChat, XSUM, SQuAD1.1, and 8k target tokens for ROCStory and Quora.
We use the early-stop trick according to the performance on the validation set. It usually takes less than 60k steps on SQuAD1.1, Quora, and PersonaChat, and 100k steps on XSUM and ROCStory.
We tune the glancing ratio from \{0.3, 0.5\}, and learning rate from \{1e-5, 2e-5, 5e-5, 1e-4, 2e-4\}. We evaluate the model every 5k steps on the validation set and obtain the final model by averaging the five best checkpoints.

\begin{table}[t]

\begin{center}
\begin{small}
\resizebox{1\linewidth}{!}{
\setlength{\tabcolsep}{1mm}{
\begin{tabular}{llcccc}
\toprule
\bf Dataset & \bf Task & \bf \# Samples & \bf Length \\
\hline
SQuAD1.1$^\spadesuit$ & Question Generation & 75k/10k/12k & 149.4/11.5 \\
XSUM$^\spadesuit$ & Summarization & 204k/11k/11k & 358.5/21.2 \\
Quora$^\heartsuit$ & Paraphrase Generation & 138k/5k/4k & 11.5/11.5 \\
PersonaChat$^\spadesuit$ & Dialog Generation & 122k/15k/14k & 120.8/11.8 \\
ROCStory$^\clubsuit$ & Story Generation & 88k/5k/5k & 9.2/41.6 \\
\bottomrule
\end{tabular}

}
}
\end{small}
\end{center}
\vspace{-0.5em}
\caption{Dataset statistics. \textbf{\# Samples} shows the number of samples in training/validation/test set. \textbf{Length} shows the average length of input/target. We use the processed datasets and evaluation metrics from $\spadesuit$~\citet{glge2021liu}, $\heartsuit$~\citet{mist2021jiang}, $\clubsuit$~\citet{comstory2020guan}.}
\label{tab:datasets}
\vspace{-0.5em}
\end{table}

\begin{table*}[t]
\begin{center}
\begin{small}
\resizebox{1\linewidth}{!}{
\setlength{\tabcolsep}{2.3mm}{

\begin{tabular}{l>{\hspace*{-5mm}}cr@{}lr@{}lr@{}lr@{}lr@{}lr@{}lr@{}lr@{}lr@{}l>{\columncolor{LightCyan}}c>{\hspace*{1mm}}r>{\hspace*{-2mm}}l>{\hspace*{1mm}}r>{\hspace*{-2mm}}l}
\toprule
\multirow{2}{*}{\bf Model} & \bf Pre- & \multicolumn{6}{c}{\bf SQuAD1.1} & \multicolumn{6}{c}{\bf XSUM} & \multicolumn{6}{c}{\bf Quora} & \multicolumn{1}{c}{\multirow{2}{*}{\bf Avg.}} & \multicolumn{2}{c}{\bf Latency} & \multicolumn{2}{c}{\bf Throughput}\\
\cmidrule(lr){3-8} \cmidrule(lr){9-14} \cmidrule(lr){15-20}
 & \bf trained? & \multicolumn{2}{c}{R-L} & \multicolumn{2}{c}{B-4} & \multicolumn{2}{c}{MTR} & \multicolumn{2}{c}{R-1} & \multicolumn{2}{c}{R-2} & \multicolumn{2}{c}{R-L} & \multicolumn{2}{c}{B-1} & \multicolumn{2}{c}{B-4} & \multicolumn{2}{c}{MTR} & \multicolumn{1}{c}{} & \multicolumn{2}{c}{ms/sample} & \multicolumn{2}{c}{samples/s}\\ \specialrule{.4pt}{0.4pt}{0.2pt}
\multicolumn{25}{c}{\textit{\footnotesize{Autoregressive Text Generation Models}}} \\
\specialrule{.4pt}{0.2pt}{1pt}
Transformer & N & 29 & .43 & 4 & .61 & 9 & .86 & 30 & .66 & 10 & .80 & 24 & .48 & 58 & .57 & 30 & .14 & 31 & .79 & 25.59 & \multicolumn{2}{c}{$-$} & \multicolumn{2}{c}{$-$}\\
MASS & Y & 49 & .48 & 20 & .16 & 24 & .41 & 39 & .70 & 17 & .24 & 31 & .91 & 60 & .56 & 32 & .39 & 32 & .92 & 34.31 & 353 & (1.0x) & 12 & (1.0x)\\
BART & Y & 42 & .55 & 17 & .06 & 23 & .19 & 38 & .79 & 16 & .16 & 30 & .61 & 61 & .56 & 31 & .57 & 32 & .42 & 32.66 & \multicolumn{2}{c}{$-$} & \multicolumn{2}{c}{$-$}\\
ProphetNet & Y & 48 & .00 & 19 & .58 & 23 & .94 & \underline{39} & \underline{.89} & 17 & .12 & 32 & .07 & 62 & .59 & \underline{33} & \underline{.80} & \underline{33} & \underline{.95} & 34.55 & \multicolumn{2}{c}{$-$} & \multicolumn{2}{c}{$-$}\\
\specialrule{.4pt}{1pt}{0.2pt}
\multicolumn{25}{c}{\textit{\footnotesize{Non-autoregressive Text Generation Models}}} \\
\specialrule{.4pt}{0.2pt}{1pt}
Vanilla NAT & N & 31 & .51 & 2 & .46 & 8 & .86 & 24 & .04 & 3 & .88 & 20 & .32 & 39 & .85 & 9 & .33 & 18 & .90 & 17.68 & \multicolumn{2}{c}{$-$} & \multicolumn{2}{c}{$-$}\\
GLAT+CTC & N & 30 & .31 & 3 & .21 & 10 & .21 & 31 & .34 & 9 & .06 & 24 & .68 & 58 & .96 & 26 & .67 & 30 & .55 & 25.00 & 24 & (14.7x) & 267 & (21.5x)\\
DSLP+CTC & N & 28 & .70 & 3 & .00 & 10 & .59 & 28 & .75 & 7 & .35 & 22 & .73 & 61 & .12 & 29 & .70 & 32 & .37 & 24.92 & 24 & (14.7x) & 265 & (21.4x)\\
LatentGLAT & N & 28 & .28 & 2 & .38 & 10 & .43 & 28 & .44 & 7 & .00 & 22 & .66 & 59 & .78 & 28 & .30 & 31 & .26 & 24.28 & 28 & (12.8x) & 334 & (27.0x)\\
BANG & Y & 44 & .07 & 12 & .75 & 18 & .99 & 32 & .59 & 8 & .98 & 27 & .41 & 55 & .18 & 24 & .97 & 25 & .99 & 27.88 & \bf 18 & \bf (19.6x) & \bf 360 & \bf (29.0x)\\
MIST & Y & 47 & .13 & 16 & .00 & 21 & .10 & 34 & .63 & 11 & .29 & 28 & .70 & 59 & .65 & 29 & .00 & 31 & .56 & 31.01 & 22 & (15.9x) & 159 & (12.8x)\\
\specialrule{.4pt}{0.2pt}{0.6pt}
PreDAT (Ours) & Y & 49 & .78 & 21 & .74 & 24 & .58 & 38 & .80 & 16 & .07 & 31 & .78 & \bf \underline{62} & \bf \underline{.63} & 32 & .59 & 33 & .37 & 34.59 & 26 & (13.8x) & 278 & (22.5x)\\
\quad w/ BeamSearch & Y & \bf \underline{50} & \bf \underline{.41} & \bf \underline{22} & \bf \underline{.66} & \bf \underline{25} & \bf \underline{.11} & \bf 39 & \bf .79 & \bf \underline{17} & \bf \underline{.38} & \bf \underline{32} & \bf \underline{.71} & 62 & .62 & \bf 33 & \bf .18 & \bf 33 & \bf .52 & \bf \underline{35.26} & 63 & (5.7x) & 214 & (17.3x)\\
\quad w/o Pre-training & N & 30 & .11 & 3 & .30 & 10 & .32 & 32 & .56 & 11 & .17 & 26 & .21 & 59 & .82 & 28 & .17 & 31 & .10 & 25.86 & 25 & (14.3x) & 272 & (21.9x)\\
\bottomrule
\end{tabular}

}
}
\end{small}
\end{center}
\vspace{-0.5em}
\caption{Performance on closed-ended text generation datasets. \textbf{Bold} and \underline{underlined} values indicate the best methods in NAR models and all models, respectively. \textbf{Latency} measures the average time of processing samples with a batch size of 1, and \textbf{Throughput} measures the speed of processing samples with a large batch size (tuned to maximize the throughput), which are evaluated on the test set of XSUM. The metrics include ROUGE-1/2/L (R-1/2/L), BLEU-1/4 (B-1/4), and METEOR (MTR).}
\label{tab:main_result}
\vspace{-1em}
\end{table*}

\vspace{0.2em}
\noindent\textbf{Inference}\ \ 
We utilize lookahead decoding (default unless otherwise specified) and beamsearch~\cite{dat2022huang} to decode a sequence from predicted DAG. We use a beam size of 200 and incorporate a 5-gram LM in the beam search.
For open-ended generation, we further employ Nucleus sampling \cite{topp2020holtzman}.

For these three decoding strategies, we prevent any repeated tri-gram in expanding the decoding path on the DAG, which is inspired by a similar strategy used in autoregressive decoding \cite{norepeatngram2018paulus}. Moreover, we also prevent consecutive uni-gram and bi-gram repetitions, which are common errors in \methodname's outputs.

\subsection{Experiment Settings}
\label{sec:settings}

\textbf{Datasets and Metrics}\quad
We utilize five datasets: SQuAD1.1 {\cite{squad2016rajpurkar}}, XSUM {\cite{xsum2018narayan}}, Quora\footnote{\url{https://quoradata.quora.com/First-Quora-Dataset-Release-Question-Pairs}}, PersonaChat {\cite{personachat2018zhang}}, and ROCStory {\cite{rocstory2016mostafazadeh}}.
We use the processed datasets and the evaluation metrics from previous work, as shown in Table \ref{tab:datasets}.
Note that we use corpus BLEU \cite{bleu2002papineni} on all datasets because the sentence BLEU may unreasonably prefer very long outputs due to the smoothing method.\footnotemark
\footnotetext{Some previous work~\cite{glge2021liu} utilize nltk's sentence BLEU with \textsf{SmoothingFunction().method7}.}

To evaluate the decoding speedup, we use two metrics:
latency measures the average time of processing a single sample,
and throughput measures the average speed in processing the whole test set, where we tune the batch size to maximize the throughput.
All models except MIST are implemented with Fairseq~\cite{fairseq2019ott} + Apex, where MIST is implemented with Huggingface's Transformers~\cite{huggingfacetrans2019wolf}. For the beam search algorithm on DAG, we adopt the C++ implementation provided by \citet{dat2022huang}. The C++ optimization only affects the extra decoding step on the CPU, but does not speedup the transformer model.
All results of speed are evaluated on a workstation with an Nvidia V100-32G GPU and 2 Intel Xeon Gold 6226R CPUs with 32 cores.

\begin{figure*}[!t]
\begin{minipage}[b]{.70\textwidth}

\begin{center}
\begin{small}
\resizebox{1.0\linewidth}{!}{
\setlength{\tabcolsep}{1.5mm}{

\begin{tabular}{l>{\hspace*{-5mm}}cr@{}lr@{}lr@{}lr@{}lr@{}lr@{}lr@{}l>{\hspace*{1mm}}r>{\hspace*{-2mm}}l>{\hspace*{1mm}}r>{\hspace*{-2mm}}l}
\toprule
\multirow{2}{*}{\bf Model} & \bf Pre- & \multicolumn{8}{c}{\bf PersonaChat} & \multicolumn{6}{c}{\bf ROCStory} & \multicolumn{2}{c}{\bf Latency} & \multicolumn{2}{c}{\bf Throughput}\\
\cmidrule(lr){3-10} \cmidrule(lr){11-16}
 & \bf trained? & \multicolumn{2}{c}{B-1} & \multicolumn{2}{c}{B-2} & \multicolumn{2}{c}{D-1} & \multicolumn{2}{c}{D-2} & \multicolumn{2}{c}{B-1} & \multicolumn{2}{c}{B-2} & \multicolumn{2}{c}{D-4} & \multicolumn{2}{c}{ms/sample} & \multicolumn{2}{c}{samples/s}\\ \specialrule{.4pt}{0.4pt}{0.2pt}
\multicolumn{20}{c}{\textit{\footnotesize{Autoregressive Text Generation Models}}} \\
\specialrule{.4pt}{0.2pt}{1pt}
Transformer & N & 18 & .37 & 8 & .07 & 1 & .43 & 10 & .04 & 30 & .68 & 14 & .67 & 35 & .18 & 168 & (1.1x) & 28 & (1.1x)\\
MASS & Y & 26 & .82 & 14 & .70 & 1 & .20 & 7 & .58 & 35 & .02 & 16 & .96 & 51 & .20 & 180 & (1.0x) & 25 & (1.0x)\\
\quad w/ Sampling & Y & 23 & .90 & 12 & .13 & 1 & .85 & 13 & .09 & 32 & .56 & 14 & .97 & 73 & .72 & 130 & (1.4x) & 77 & (3.0x)\\
BART & Y & 26 & .84 & 14 & .69 & 1 & .39 & 8 & .85 & \underline{35} & \underline{.45} & 17 & .22 & 49 & .03 & 199 & (0.9x) & 23 & (0.9x)\\
\quad w/ Sampling & Y & 24 & .00 & 12 & .31 & 1 & .97 & 14 & .50 & 33 & .95 & 15 & .28 & 73 & .62 & 143 & (1.3x) & 69 & (2.7x)\\
\specialrule{.4pt}{1pt}{0.2pt}
\multicolumn{20}{c}{\textit{\footnotesize{Non-autoregressive Text Generation Models}}} \\
\specialrule{.4pt}{0.2pt}{1pt}
Vanilla NAT & N & 18 & .33 & 6 & .37 & 0 & .43 & 0 & .96 & 28 & .44 & 11 & .29 & 89 & .13 & 23 & (7.8x) & \bf 703 & \bf (27.7x)\\
BANG & Y & 17 & .38 & 7 & .33 & \bf \underline{2} & \bf \underline{.12} & \bf \underline{23} & \bf \underline{.02} & 29 & .38 & 11 & .78 & \bf \underline{92} & \bf \underline{.10} & \bf 18 & \bf (10.1x) & 649 & (25.6x)\\
MIST & Y & 18 & .55 & 8 & .86 & 0 & .54 & 2 & .56 & 23 & .57 & 9 & .09 & 8 & .15 & 25 & (7.3x) & 330 & (13.0x)\\
\specialrule{.4pt}{0.2pt}{0.6pt}
PreDAT (Ours) & Y & 27 & .06 & 15 & .05 & 1 & .33 & 8 & .31 & 34 & .11 & 17 & .17 & 57 & .50 & 24 & (7.6x) & 507 & (20.0x)\\
\quad w/ Sampling & Y & 24 & .23 & 12 & .29 & 1 & .77 & 15 & .62 & 32 & .52 & 15 & .61 & 74 & .37 & 24 & (7.4x) & 514 & (20.3x)\\
\quad w/ BeamSearch & Y & \bf \underline{27} & \bf \underline{.31} & \bf \underline{15} & \bf \underline{.39} & 1 & .15 & 6 & .30 & \bf 34 & \bf .61 & \bf \underline{17} & \bf \underline{.84} & 50 & .55 & 48 & (3.7x) & 318 & (12.6x)\\
\quad w/o Pre-training & N & 21 & .96 & 10 & .38 & 0 & .52 & 3 & .29 & 31 & .81 & 15 & .41 & 52 & .97 & 25 & (7.2x) & 562 & (22.2x)\\
\bottomrule
\end{tabular}

}
}
\end{small}
\end{center}
\vspace{-0.5em}
\captionof{table}{Performance on open-ended text generation datasets. \textbf{Latency} and \textbf{Throughput} are evaluated on the test set of PersonaChat. Average scores are not shown because they cannot reflect the trade-off between quality and diversity. We utilize corpus BLEU on all datasets, whose values may be different from some previous results utilizing sentence BLEU~\cite{glge2021liu}. The metrics include BLEU-1/2 (B-1/2) and Distinct-1/2/4 (D-1/2/4).}
\label{tab:main_result2}

\end{minipage}
\hfill
\begin{minipage}[b]{.27\textwidth}

\centering
\vspace{-0em}
\includegraphics[width=0.9\textwidth]{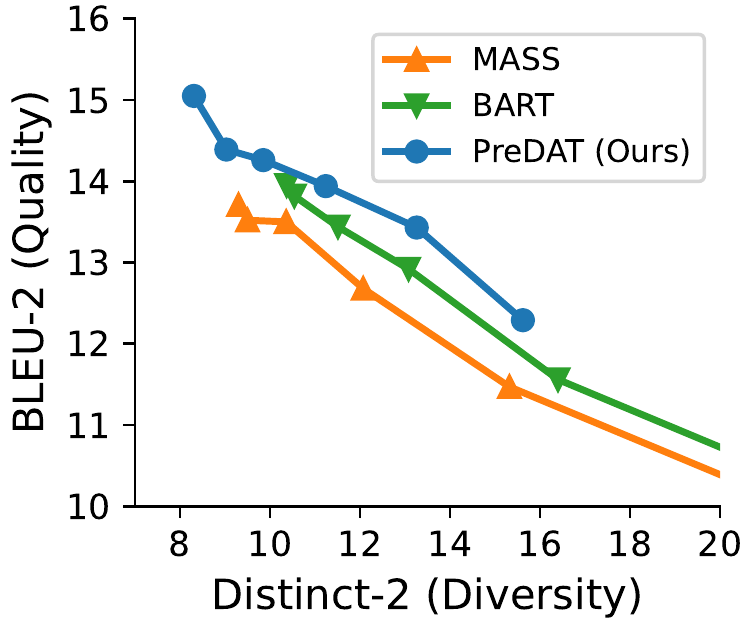}
\vspace{-0em}
\caption{Trade-off curves of quality and diversity on PersonaChat. All models use Nucleus sampling with $p=0.9$ and temperature $\tau$ from $\{0, 0.2, 0.4, 0.6, 0.8, 1\}$.}
 \label{fig:sampling}
\end{minipage}
\vspace{-0.5em}
\end{figure*}

\vspace{0.2em}
\noindent\textbf{Baselines}\quad
Our baselines include autoregressive Transformer~\cite{transformer2017vaswani}, pre-trained AR models (MASS, \citealp{mass2019song}; BART, \citealp{bart2020lewis}; ProphetNet, \citealp{prophetnet2020qi}), non-pretrained NAR models (Vanilla NAT, \citealp{nat2018gu}; GLAT+CTC, \citealp{glat2021qian}; DSLP+CTC, \citealp{dslp2021huang}; LatentGLAT, \citealp{latentglat2022bao}), pre-trained NAR models (BANG, \citealp{bang2021qi}; MIST, \citealp{mist2021jiang}). All these baselines have the same number of layers and hidden sizes as our \methodname, except that LatentGLAT utilizes a 4-layer latent predictor and a 4-layer decoder based on the original implementation.
Note that CTC-based models also require an up-sampling strategy, so we add a length predictor following the description of Sec.\ref{sec:finetune}. Their up-sampling ratio is sampled from $[1.5, 2]$ in training and tuned on the validation set in inference.
For AR baselines, unless otherwise specified, we use BeamSearch with a beam size of 5 and the tri-gram repetition prevention trick~\cite{norepeatngram2018paulus}, and tune the length penalty on the validation set.
For NAR baselines, we use greedy decoding and further remove consecutive repeated tokens after generation~\cite{hint2019li}.
Some results are collected from \citet{glge2021liu, bang2021qi, mist2021jiang}.

\subsection{Automatic Evaluation}
\label{sec:automatic}

\textbf{Closed-Ended Text Generation}\quad
We first test \methodname on three closed-ended text generation tasks, including question generation, summarization, and paraphrase generation.
Closed-ended text generation tasks usually have strict semantic constraints on the outputs, aiming to test the model's ability to extract and organize information.

As shown in Table \ref{tab:main_result}, \methodname achieves surprisingly good results in both speed and quality. We highlight our advantages as follows:

\vspace{-0.5em}
\begin{itemize}[leftmargin=1em]
    \setlength{\itemsep}{0ex}{
    \setlength{\parskip}{2px}{
    \item \methodname remarkably imrpoves the quality of NAR generation. Compared with previous pretrained NAR models, \methodname brings large improvement (+4.2 scores on average) due to our Double-Source Text Infilling pre-training and the DAT architecture. Moreover, \methodname even outperforms the best AR baseline by 0.7 scores. To our knowledge, it is the first time that an NAR model achieves comparable and even stronger performance than AR models in n-gram-based metrics on a wide range of text generation tasks.
    \item \methodname is highly efficient. Although our model is slightly slower than previous NAR models due to a longer sequence prediction, it still achieves a speedup of 5$\sim$14 times in latency and 17$\sim$23 times in throughput compared with AR generation. It verifies that \methodname can largely reduce computing consumption in decoding, showing its potential for real-time applications.
    }
    }
\end{itemize}
\vspace{-0em}

\noindent\textbf{Open-Ended Text Generation}\quad
We further test \methodname on two open-ended text generation tasks, dialog generation and story generation.
Open-ended text generation tasks encourage the model to produce novel and diverse outputs, where sampling decoding methods are commonly adopted to promote generation diversity.

Therefore, in addition to lookahead decoding and beamsearch, we also introduce Nucleus sampling \cite{topp2020holtzman}.
Specifically, we set $p=0.9$ and the temperature $\tau=1$ for \methodname. For MASS and BART, we also use $p=0.9$, but $\tau=0.8$ on PersonaChat and $\tau=0.7$ on ROCStory to achieve similar diversity as \methodname.%

We present the evaluation results in Table \ref{tab:main_result2} and the trade-off of quality and diversity by tuning the temperature in Fig.\ref{fig:sampling}.
Generally, the comparison of quality metrics is similar to closed-ended generation: \methodname largely outperforms NAR baselines and achieves comparable BLEU scores to AR models. 
Moreover, we highlight two findings:

\vspace{-0.5em}
\begin{itemize}[leftmargin=1em]
    \setlength{\itemsep}{0ex}{
    \setlength{\parskip}{2px}{
    \item \methodname generates plausible outputs in open-ended tasks while previous NAR models cannot. Open-ended generation tasks usually have targets with diverse expressions, which worsens the multi-modality problem and seriously degrades the NAR generation quality. Specifically, MIST shows very low diversity because it generates a lot of repetitions, and BANG shows very high diversity because it introduces many incomprehensible n-grams. In contrast, \methodname has a reasonable quality-diversity trade-off, showing its ability to address the serious challenges brought by the multi-modality problem.
    
    \item \methodname achieves a flexible quality and diversity trade-off.
    As shown in Fig.\ref{fig:sampling}, \methodname is slightly better than two AR baselines w.r.t. the trade-off curves by tuning the decoding temperature. It demonstrates that \methodname can meet the diversity requirement of open-ended text generation, verifying its generality in text generation. %
    }
    }
\end{itemize}
\vspace{-0.5em}

\subsection{Ablation Study}
\label{sec:ablation}

In this section, we conduct ablation studies to reveal how our designs contribute to the results.

\begin{figure}
\centering
\vspace{-0em}
\includegraphics[width=0.6\linewidth]{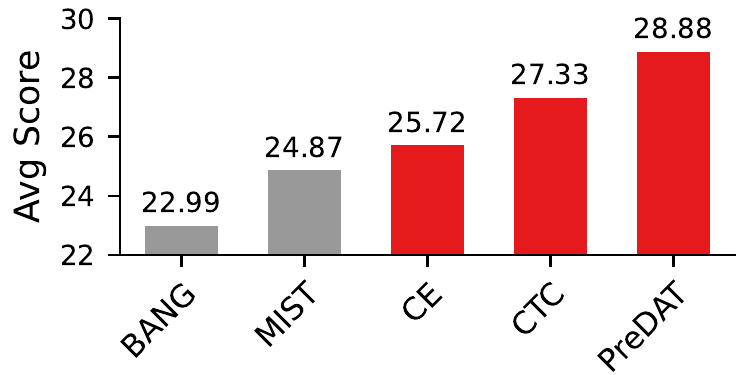}
\vspace{-1em}
\caption{Average performance of previous baselines (\textcolor{gray}{Gray}) and NAR models pre-trained by our proposed task with different loss functions (\textcolor{red}{Red}). The shown scores are the average of automatic metrics on XSUM.}
\vspace{-1em}
\label{fig:lossfunc}
\end{figure}

\vspace{0.2em}
\noindent\textbf{Loss Function}\quad
In \methodname, we utilize the DAT loss to alleviate the multi-modality problem, which plays an important role in the pre-training.
Notably, our pre-training task can be combined with other NAR losses, so we compare the DAT loss against CTC \cite{ctc2006graves, ctc2018libovicky} and the token-level cross-entropy loss (CE).

Specifically, the same loss function is applied in both pre-training and fine-tuning to avoid discrepancies between the two training stages. For CTC, we randomly sample the up-sampling ratio from $[1.5, 2]$. For CE, we do not use up-sampling (i.e., $\lambda=1$) because the CE loss requires an output sequence with the same length as the target.

As shown in Fig.\ref{fig:lossfunc}, we find:
(1) It is important to incorporate alignment-based NAR losses in pre-training, where CTC and DAT losses bring substantial improvements compared with the CE loss.
(2) The NAR model pre-trained with CE still outperforms previous pre-trained NAR baselines, verifying the effectiveness of our pre-training task in preserving sentence-level consistency and promoting bidirectional dependencies.

\begin{figure}
\centering
\vspace{-0em}
\includegraphics[width=0.95\linewidth]{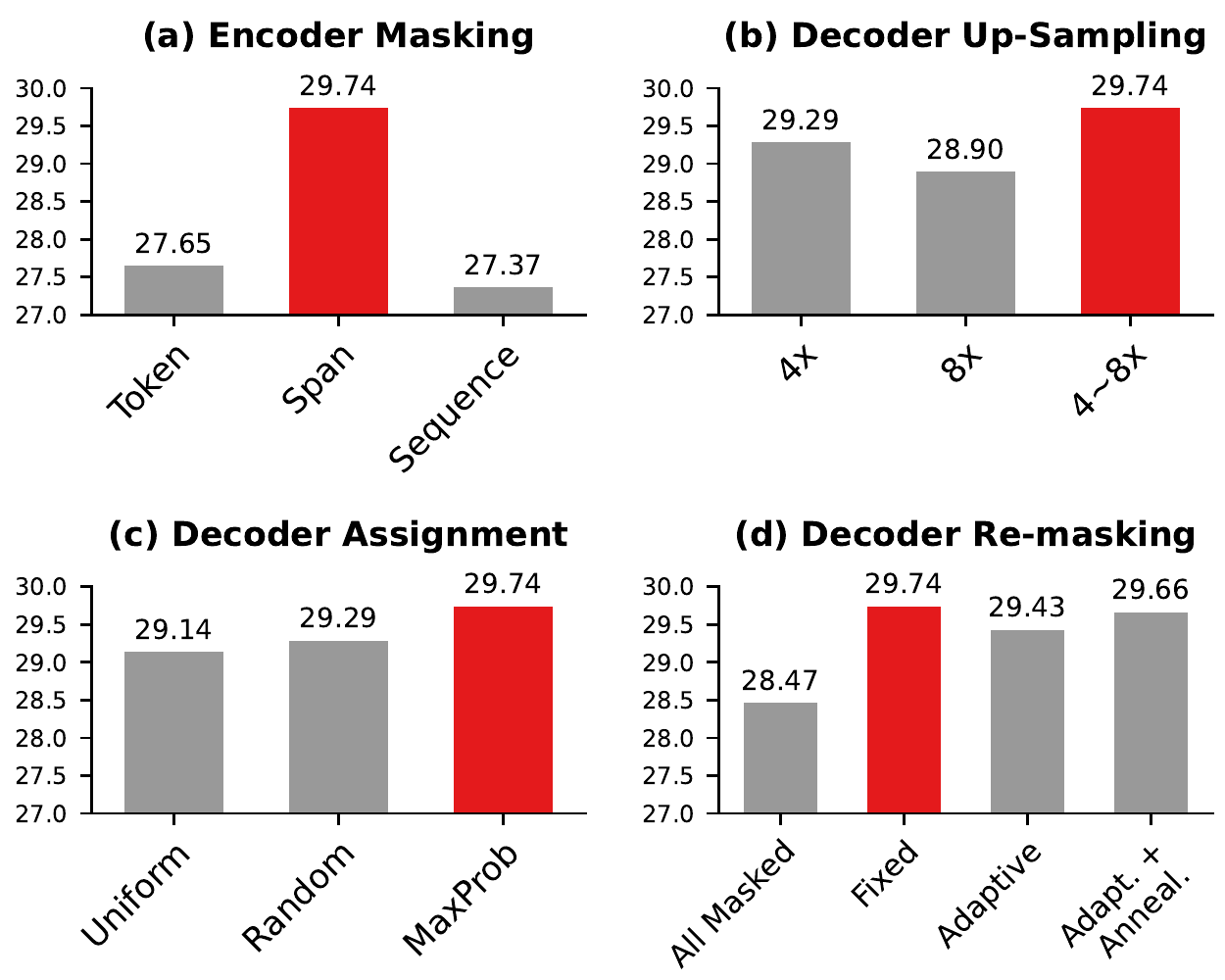}
\vspace{-1em}
\caption{Comparisons of pre-training strategies by the average validation score on SQuAD1.1 and XSUM. All models are pre-trained for 100k steps for energy saving. The strategies in our final model are marked in \textcolor{red}{Red}.}
\vspace{-1.2em}
\label{fig:ablation}
\end{figure}

\vspace{0.2em}
\noindent\textbf{Pre-training Strategy}\ \ 
Our proposed pre-training task includes several strategies for constructing the training data pair.
To evaluate the effects of these strategies, we design four groups of comparisons as follows, whose results are shown in Fig.\ref{fig:ablation}.

(a) Stage 1: Encoder Masking.
Besides \textbf{Span} masking, we use two other strategies including \textbf{Token} masking that independently samples masked positions \cite{bert2019devlin}, and \textbf{Sequence} masking that masks a single consecutive sequence. All strategies mask the same ratio of tokens.
We conclude that the masked spans should not be too short (about 1$\sim$3 tokens in token masking) or too long (about 90 tokens in sequence masking), which prevents the NAR model from learning prediction consistency or make the prediction too difficult.

(b) Stage 2, Step 1: Up-sample Ratios.
We compare the random sampling ratio (\textbf{4$\sim$8x}) against fixed up-sampling ratios (\textbf{4x} and \textbf{8x}).
We find that random up-sampling can diversify the DAG structure, which works as a data augmentation method and thus benefits the downstream performance.

(c) Stage 2, Step 2: Assignment Strategies. 
Besides the proposed assignment strategy according to the path probability (\textbf{MaxProb}), we use \textbf{Uniform} and \textbf{Random} assignment that assigns the target into the decoder input uniformly or randomly.
We find the MaxProb assignment can better determine the lengths of each masked segment according to the model's own prediction, leading to slightly better results than the other strategies.

(d) Stage 2, Step 3: Re-masking Strategies. 
Besides the \textbf{Fixed} masking strategy, we also try \textbf{Adaptive} and \textbf{Adaptive + Annealing} masking strategies proposed by \citet{glat2021qian}, where they adjust the masking ratio by the difficulties of the sample.
It shows that these strategies have similar performance, outperforming the fully masked decoder input (\textbf{All Masked}),
which verifies the importance of introducing information in the decoder input for bidirectional dependency modelling.
However, the adaptive masking is less effective in pre-training, so we use the fixed masking ratio for simplicity.

\vspace{0.2em}
\noindent\textbf{Up-sampling Ratio in Fine-tuning}\quad
As described in Sec.\ref{sec:finetune}, we obtain the decoder input length in fine-tuning by up-sampling.
To investigate how the up-sampling strategies affect performance, we evaluate different combinations of up-sampling ratios in pre-training and fine-tuning.

\begin{figure}
\centering
\vspace{-0em}
\includegraphics[width=0.82\linewidth]{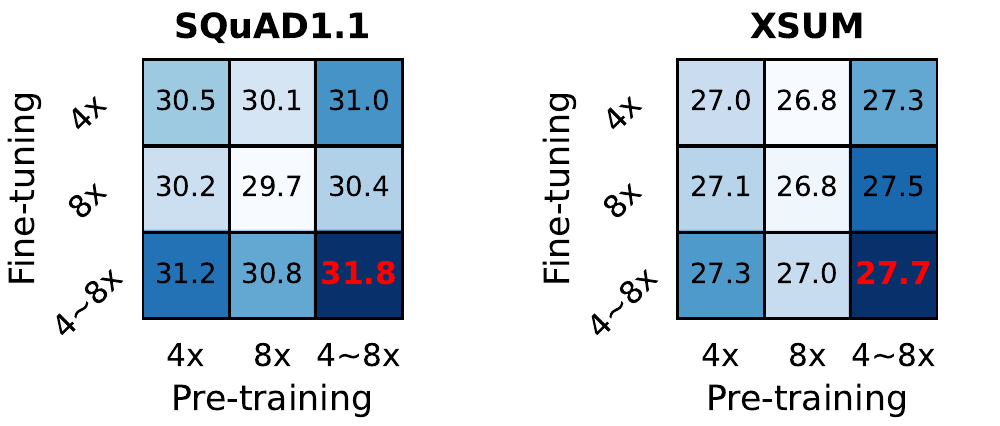}
\vspace{-1em}
\caption{Validation performance under various combinations of up-sampling strategies in pre-training and fine-tuning. The shown score is the average of automatic metrics. 4x and 8x indicates fixed up-sampling ratios, and 4$\sim$8x indicates random ratios sampled from $[4, 8]$. All models are pre-trained for only 100k steps.}
\vspace{-0.5em}
 \label{fig:upsample}
\end{figure}

As shown in Fig.\ref{fig:upsample}, random up-sampling always benefits the performance in pre-training and fine-tuning, together bringing an improvement of about 1.2 scores.
It indicates that varying the DAG size is an important trick in training \methodname.
Moreover, the up-sampling ratios in pre-training and fine-tuning do not need to be the same, which can be helpful if smaller DAG sizes are preferred in downstream tasks due to limited memory budget.

\begin{figure}[t]
\centering
\vspace{-0em}
\includegraphics[width=0.9\linewidth]{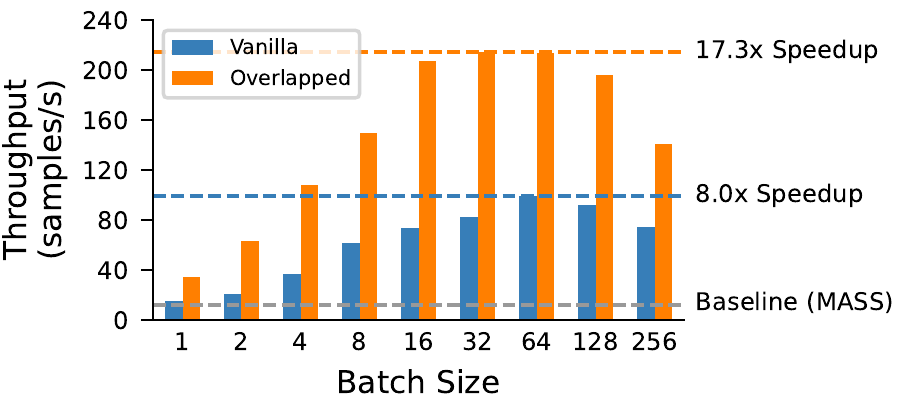}
\vspace{-0.8em}
\caption{Throughput speedups with the vanilla and overlapped decoding on the test set of XSUM.}
\vspace{-1.1em}
\label{fig:speedup}
\end{figure}

\vspace{0.2em}
\noindent\textbf{Overlapped Decoding}\ \ 
Overlapped decoding aims to improve the decoding throughput by overlapping the execution of DAG prediction and beam search decoding. To verify its effectiveness, we evaluate the speedup with various batch sizes on XSUM.

As shown in Fig.\ref{fig:speedup}, our overlapped decoding brings a 17.3x speedup with a batch size of 32, largely outperforming the vanilla one. We also note that throughput starts to decline as batch size increases, possibly because the introduced paddings increase the consumption of invalid computations.

\subsection{Analysis}
\label{sec:analysis}

In this section, we investigate the reasons why \methodname achieves better automatic scores than pre-trained AR baselines, which may provide some insights for future NAR generation studies.

\begin{figure}
\centering
\vspace{-0em}
\includegraphics[width=\linewidth]{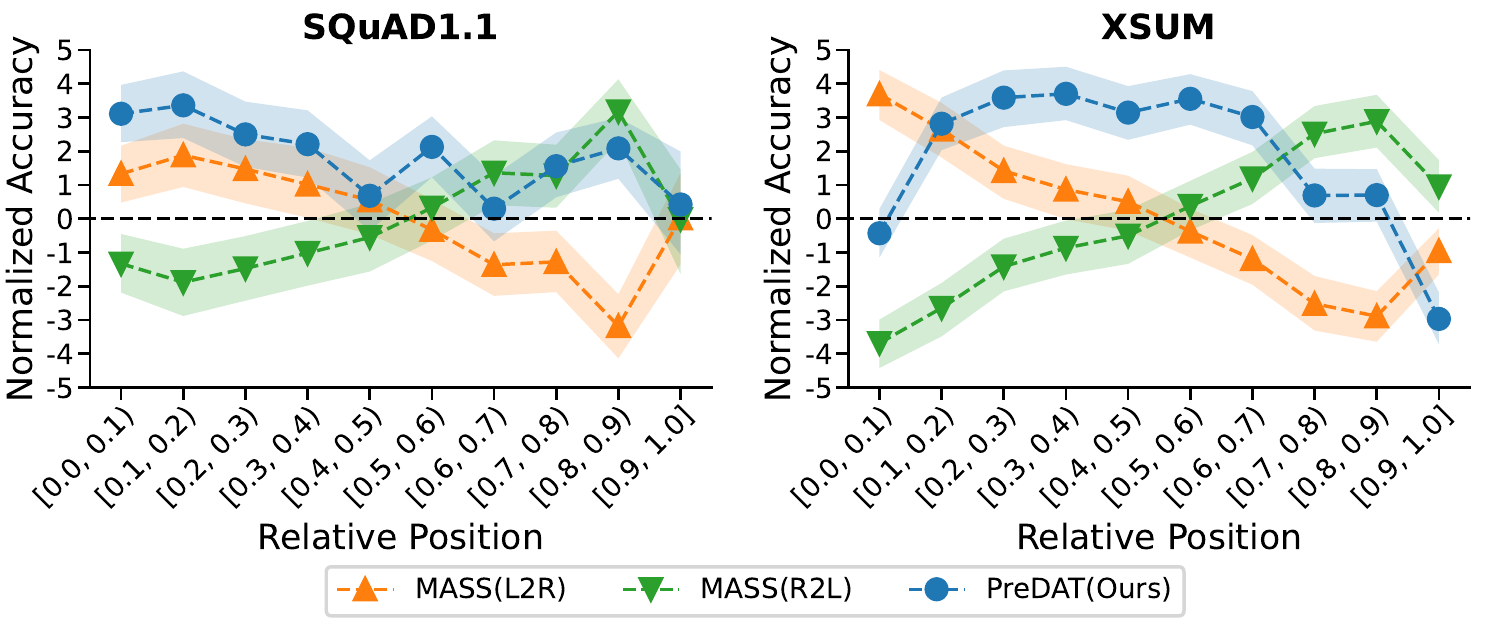}
\vspace{-2em}
\caption{Normalized prediction accuracy $\Delta \text{Acc}(D)$ bucketed by relative positions. The shaded area is 95\% confidence interval by bootstrap sampling~\cite{bootstrap1993efron}. L2R: left-to-right, R2L: right-to-left.}
\vspace{-1em}
 \label{fig:wordacc}
\end{figure}

\vspace{0.2em}
\noindent\textbf{\methodname alleviates error accumulation.}\quad
Error accumulation is a major concern of autoregressive generation~\cite{bengio2015schedule, ranzato2016seqlevel, arora2022exposurebias}, where a prediction error may be propagated into later decoding steps, leading to low-quality generated sentences.
 In contrast, NAR models naturally avoid the problem due to their unbiased prediction order.

To verify that \methodname has advantages in tackling error accumulation, we compare \methodname against two AR models with different decoding orders, a left-to-right (L2R) one and a right-to-left (R2L) one.
Specifically, we fine-tune MASS on the downstream datasets using the two generation orders. We find that MASS still performs well in right-to-left decoding, with a performance drop of less than 0.5 scores.
Then we calculate the average token prediction accuracy bucketed by the relative position, formally defined as

\vspace{-1em}
{
\small
\begin{gather}
    \text{Acc}(D) = \text{Average}( \mathbb{I}( \hat{Y}^{(i)}_j \in Y^{(i)}) )\notag \\
    \quad\quad \text{for}\ \  1\leq i \leq N, 1 \leq j \leq |\hat{Y}^{(i)}|, \frac{j}{|\hat{Y}^{(i)}| + 1} \in D, \notag
\end{gather}
}
\vspace{-1em}

\noindent where $\text{Acc}(D)$ is the average prediction accuracy on the interval $D$, ${Y}^{(i)}$ is the i-th sample in the test set, $\hat{Y}^{(i)}$ is the i-th model outputs, and $j$ indicates the position.
Moreover, since prediction difficulties vary with the positions (e.g., the last token is always punctuation), we utilize a normalized accuracy:

\vspace{-1em}
{
\small
\begin{gather}
\Delta \text{Acc}(D) = \text{Acc}(D) - \frac{\text{Acc}_{\text{L2R}}(D) + \text{Acc}_{\text{R2L}}(D)}{2}, \notag
\end{gather}
}
\vspace{-1em}

\noindent where $\text{Acc}_{\text{L2R}}(D)$ and $\text{Acc}_{\text{R2L}}(D)$ indicates the prediction accuracy of L2R and R2L MASS.

As shown in Fig.\ref{fig:wordacc}, we find that MASS has a strong tendency that it predicts earlier generated tokens more accurately than later generated ones, which applies to both left-to-right and right-to-left models. In contrast, our \methodname shows no significant preference for any positions because it predicts all tokens simultaneously,
which reveals the advantages of unbiased prediction order in NAR generation models.

\vspace{0.2em}
\noindent\textbf{\methodname improves the relevance to the input.}\quad
Previous studies empirically found that AR generated texts may lose relevance to the input sentences, which is also known as hallucination~\cite{faithfulness2020maynez, hallucinationsurvey2022ji} or off-prompt errors~\cite{scarecrow2022dou}.
One explanation is that AR models may be distracted by its generated prefixes, which can be avoided in NAR generation~\cite{huang2021nast}.

To verify our hypothesis, we introduce two metrics to evaluate the relevance to inputs: Knowledge F1 \cite{knowledgef12021shuster} and PARENT-T \cite{parent2019dhingra,parentt2020wang}. Knowledge F1 measures the unigram F1 between generated sentences and the input knowledge, while PARENT-T measures n-gram entailment.
Both metrics require the extraction of knowledge pieces that should appear in the generated sentences. For simplicity, we take each sentence in the passage (of XSUM) or the persona profile (of PersonaChat) as a piece of knowledge and further filter out the stop words.

\begin{table}[!t]
\begin{center}
\begin{small}

\resizebox{0.90\linewidth}{!}{
\setlength{\tabcolsep}{1.5mm}{
\begin{tabular}{clrrrrrr}
\toprule
\multirow{2}{*}{\bf Dataset} & \multirow{2}{*}{\bf Model} & \multicolumn{3}{c}{\bf Knowledge} & \multicolumn{3}{c}{\bf PARENT-T} \\
\cmidrule(lr){3-5} \cmidrule(lr){6-8}
  &   & \multicolumn{1}{c}{P} & \multicolumn{1}{c}{R} & \multicolumn{1}{c}{F1} & \multicolumn{1}{c}{P} & \multicolumn{1}{c}{R} & \multicolumn{1}{c}{F1}            \\
    \hline
\multirow{2}{*}{XSUM} &MASS & 35.1      & 9.7    & 14.7 & 35.1      & 8.5           & 13.1          \\
& \methodname & \textbf{36.3} & \textbf{9.9}  & \textbf{14.9} & \textbf{36.4} & \textbf{8.6} & \textbf{13.3} \\
\hline
\multirow{2}{*}{PersonaChat} & MASS & 19.6      & 17.2   & 17.8 & 13.2      & \textbf{11.3} & \textbf{11.5} \\
& \methodname & \textbf{21.1} & \textbf{17.7} & \textbf{18.5} & \textbf{13.8} & 11.0         & \textbf{11.5} \\
\bottomrule
\end{tabular}
}
}

\end{small}
\end{center}
\vspace{-0.8em}
\caption{Relevance to the input on XSUM and PersonaChat. We utilize two automatic metrics, Knowledge F1 and PARENT-T. P: Precision, R: Recall.}
\vspace{-0.5em}
\label{tab:f1}
\end{table}

As shown in Table \ref{tab:f1}, \methodname achieves better precision on both datasets in using the input knowledge compared with MASS  (+1.2 on average). It indicates that \methodname is less likely to produce irrelevant keywords, justifying our hypothesis that the NAR model can better concentrate on the input.
However, we also notice that \methodname and MASS have comparable performance on recall, showing that it is still challenging to cover more keywords.%

\subsection{Manual Evaluation}
\label{sec:manual}

\begin{table}[t]
\begin{center}
\begin{small}

\resizebox{1\linewidth}{!}{
\setlength{\tabcolsep}{1mm}{

\begin{tabular}{llllcp{0.25cm}lllc}
\toprule
\multirow{2}{*}{} & \multicolumn{4}{c}{\bf Grammaticality} && \multicolumn{4}{c}{\bf Appropriateness} \\
\cmidrule(lr){2-5} \cmidrule(lr){7-10}
& Win & Tie & Lose & $\kappa$ && Win & Tie & Lose & $\kappa$ \\
\specialrule{.4pt}{1pt}{0.2pt}
\multicolumn{10}{c}{\textit{\footnotesize{Comparison against Non-autoregressive Models}}} \\
\specialrule{.4pt}{0.2pt}{1pt}
vs. BANG & 75.3**  & 12.0    & 12.7     & 0.66  && 69.8**  & 17.3    & 12.9     & 0.59  \\
vs. MIST & 66.7**  & 18.0    & 15.3     & 0.50  && 57.1**  & 26.0    & 16.9     & 0.47  \\
\specialrule{.4pt}{1pt}{0.2pt}
\multicolumn{10}{c}{\textit{\footnotesize{Comparison against Autoregressive Models}}} \\
\specialrule{.4pt}{0.2pt}{1pt}
vs. MASS & 15.1    & 47.8    & 37.1**   & 0.32  && 32.2    & 36.7    & 31.1     & 0.46 \\
\bottomrule
\end{tabular}
}
}

\end{small}
\end{center}
\vspace{-0.5em}
\caption{Manual evaluation results on SQuAD1.1. Fleiss' $\kappa$ is shown for iter-rater reliability (all are fair agreement or above). * and ** indicate p-value<0.05 and 0.01 in the sign test, respectively.}
\label{tab:human}
\vspace{-0em}
\end{table}

Although \methodname shows surprising performance in automatic evaluation, it is still questionable whether these automatic metrics are reliable when comparing AR and NAR models.
In this section, we conduct a manual evaluation that compares \methodname against pre-trained AR and NAR baselines.

\noindent\textbf{Settings}\ \ 
We compare \methodname against three baselines, two NAR models (BANG and MIST) and an AR model (MASS). We randomly selected 150 samples in SQuAD1.1, accounting for 600 generated sentences for the four models. For each sample, three annotators were asked to rank the outputs from two dimensions:
\textbf{grammaticality} measures whether the output contains any grammatical errors, and \textbf{appropriateness} measures whether the output is reasonable for the given context.

\vspace{0.2em}
\noindent\textbf{Results}\quad
The results are shown in Table \ref{tab:human}, where we highlight two findings:
(1) \methodname achieves a significant quality improvement over previous NAR models, where annotators highly prefer \methodname (with Win\% + Tie\% $>$ 83\%).
(2) There is still a quality gap between \methodname and the AR model. Although \methodname achieves higher word overlap in automatic evaluation, it exhibits poorer grammaticality in human ratings.
A possible reason is that \methodname preserves better relevance to the inputs, leading to the higher word overlap, however, is still weaker than AR models in preserving the consistency among generated tokens.

\begin{figure}[t]
\centering
\vspace{-0em}
\includegraphics[width=0.85\linewidth]{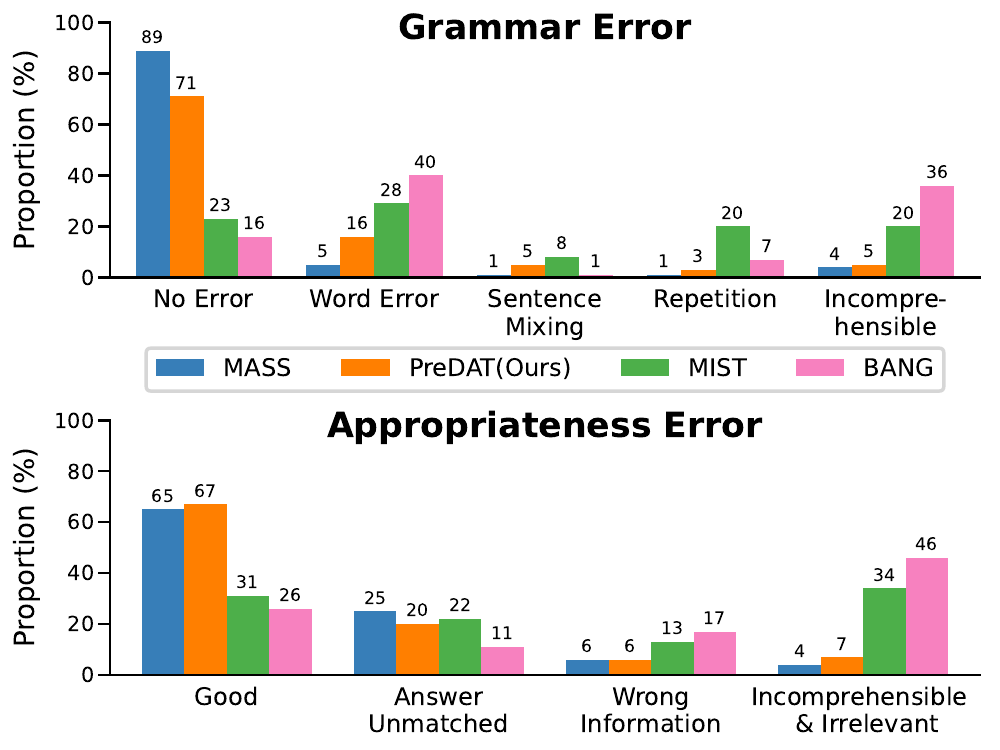}
\vspace{-1em}
\caption{Proportion of samples with different error types in terms of grammaticality and appropriateness on SQuAD1.1. \textit{Word Error}: containing less than two wrong/missing/redudant tokens. \textit{Sentence Mixing}: can be splitted into two (nearly) grammatically correct sentences with a shared fragment. \textit{Answer Unmatched}: the generated question is not matched with the given answer. \textit{Wrong Information}: using incorrect or unmentioned information in the passage.}
\vspace{-0.8em}
\label{fig:errortype}
\end{figure}

\begin{table*}[!t]
\begin{center}
\begin{footnotesize}

\resizebox{0.9\linewidth}{!}{
\setlength{\tabcolsep}{3mm}{

\begin{tabular}{ll}
\hline
\multicolumn{2}{c}{\bf SQuAD1.1} \\
\hline
\textbf{Passage:} & \multicolumn{1}{p{12.0cm}}{(74 words omitted) ... JFK and \textgood{Newark Liberty} were the busiest and \textgood{fourth busiest U.S. gateways for international air passengers}, respectively, \textgood{in 2012} ... (72 words omitted)} \\
\textbf{Answer:} & Newark Liberty International Airport \\
\hdashline
\textbf{BANG} & what airport \texterr{in busiest airport in the u} . \\
\textbf{MIST} & \multicolumn{1}{p{12.0cm}}{what is john f . kennedy international airport \texterr{john f busiest international airport and laguardia} ?} \\
\textbf{MASS} & what is the name of \textunrel{the busiest airport in new york} ? \\
\textbf{\methodname (Ours)} & \multicolumn{1}{p{12.0cm}}{what is the name of \textgood{the fourth busiest airport for international air passengers in 2012 ?}} \\
\hline
\textbf{Passage:} & \multicolumn{1}{p{12.0cm}}{(102 words omitted) ... \textgood{The FDIC guarantees the funds of all insured accounts} up to US \$ 100 , 000 ... (72 words omitted)} \\
\textbf{Answer:} & US \$ 100 , 000 \\
\hdashline
\textbf{BANG} & \texterr{what} much the deposits of \texterr{deposits of allmac deposits ?} \\
\textbf{MIST} & what is the funds of all insured \texterr{ins allmac accounts to ?} \\
\textbf{MASS} & \multicolumn{1}{p{12.0cm}}{how much does \textgood{the fdic guarantee the funds of all insured accounts} ?} \\
\textbf{\methodname (Ours)} & \multicolumn{1}{p{12.0cm}}{how much \texterr{is} \textgood{the fdic guarantee the funds of all insured accounts ?}} \\
\hline
textbf{Passage:} & \multicolumn{1}{p{12.0cm}}{When one Republican presidential candidate for the 2016 election ridiculed the liberalism of "New York values" in January 2016, \textgood{Donald Trump, leading in the polls}, vigorously defended his city ... (68 words omitted)} \\
\textbf{Answer:} & Donald Trump \\
\hdashline
\textbf{BANG} & who \texterr{did} the republican \texterr{the against new values " in} \\
\textbf{MIST} & who was the leader \texterr{the "} new york in 2016 ? \\
\textbf{MASS} & who was \textunrel{the republican presidential candidate} for 2016 ? \\
\textbf{\methodname (Ours)} & who \texterr{led} \textgood{the polls} in 2016 ? \\
\hline
\hline
\hline
\multicolumn{2}{c}{\bf PersonaChat} \\
\hline
\textbf{Persona:} & \multicolumn{1}{p{12.0cm}}{(9 words omitted) ... I like to \textgood{listen to country music} ... (24 words omitted)} \\
\textbf{History:} & A: Hello I like to travel. \\
    & B: Hello, how are you tonight? I do too and love to cook. \\
    & A: I would love to see \textgood{europe}. \\
\hdashline
\textbf{BANG} & i would like \texterr{to too} but i am a \texterr{to .} \\
\textbf{MIST} & what do you do for a living . \texterr{for a living ?} \\
\textbf{MASS} & \multicolumn{1}{p{12.0cm}}{i have never been to \textgood{europe} , but i have always wanted to go to \textunrel{australia} .} \\
\textbf{\methodname (Ours)} & \multicolumn{1}{p{12.0cm}}{i would love to go to \textgood{europe} . i am \textgood{listening to country music} .} \\
\hline
\textbf{Persona:} & \multicolumn{1}{p{12.0cm}}{I am an eccentric \textgood{hair stylist} for dogs
 ... (27 words omitted)} \\
\textbf{Dialog History:} & (24 tokens omitted) ...\\
& A: I am doing wonderful, now that I avoided the mangoes. I am allergic. \\
    & B: Oh sorry to hear that I like going out with my friends. \\
\hdashline
\textbf{BANG} & i \texterr{do you like.} \\
\textbf{MIST} & what do you do for a living \texterr{for a living ?} \\
\textbf{MASS} & do you have any pets ? \textunrel{i have a dog} . \\
\textbf{\methodname (Ours)} & what do you like to do \texterr{with} fun ? i am a \textgood{hair stylist} . \\
\hline
\textbf{Persona:} & \multicolumn{1}{p{12.0cm}}{
 ... (43 words omitted)} \\
\textbf{Dialog History:} & (131 tokens omitted) ...\\
& B: I bet you can learn a lot studying ice, must be cold though. \\
    & A: It is. Some people freeze to death. \\
    & B: Yikes, too cold for me. \textgood{i will stay home with my pets!} \\
\hdashline
\textbf{BANG} & i \texterr{do you do any} \\
\textbf{MIST} & what do you do \texterr{. pets . how . you ?} \\
\textbf{MASS} & do you have \textunrel{any hobbies besides music} ? \\
\textbf{\methodname (Ours)} & \textgood{what kind of pets} do you \texterr{do} ? \\
\hline
\end{tabular}
}
}

\end{footnotesize}
\end{center}
\vspace{-0.5em}
\caption{Cases of model outputs on SQuAD1.1 and PersonaChat. Grammatical errors are marked in \texterr{red}. The phrases that are faithful to the input are marked in \textgood{bleu}, whereas the unfaithful ones are marked in \textunrel{brown}. All generated sentences are in lowercase.}
\label{tab:case}
\vspace{-0.5em}
\end{table*}

\vspace{0.2em}
\noindent\textbf{Typical Errors and Case Study}\ \ \ \ 
To better understand how \methodname makes errors, we investigate the typical errors in the generated outputs.
Specifically, we randomly chose 100 samples from SQuAD1.1, collected the outputs of the four models, and then
manually annotated the errors in these outputs.

Fig.\ref{fig:errortype} presents the proportions of error types.%
In terms of grammaticality, we find that \methodname well addresses the major problems in previous NAR models, such as incomprehensible outputs and repetitions. 
However, there are still some word errors, which affect only a small fragment of the sentence but are very obvious to human readers, leading to the unsatisfying result. We believe the problem can be alleviated by post-editing or iterative refinement, which we leave for future work.
In terms of appropriateness, \methodname has comparable performance to MASS in error distributions, showing its ability to extract and organize information to form appropriate outputs.

To support the above discussions, we show some output cases in Table \ref{tab:case}. We find that previous NAR models usually generate low-quality texts, whereas PreDAT achieves significant improvement.
Moreover, PreDAT maintains a strong relevance to the inputs, yet it can occasionally introduce grammatical errors. In contrast, MASS generates plausible outputs, but they may not always be faithful. This observation highlights the distinctive behaviors between AR and NAR models.

\section{Limitations}

Although PreDAT achieves a significant advancement in NAR generation, it still faces the following limitations:

(1) Although PreDAT achieves superior performance in automatic evaluation, it still significantly underperforms AR models in grammaticality according to human evaluation (as discussed in Sec.\ref{sec:manual}). This inconsistency can be attributed to the different biases of AR and NAR models:
AR models tend to generate fluent outputs but may sacrifice relevance to the input, while NAR models prioritize relevance but may incur grammatical errors. It is important to take the behavior into consideration when applying PreDAT to real-world applications.

(2) PreDAT may struggle with capturing long-range coherence, because NAR models are inherently weak in modeling token dependencies, and PreDAT is pre-trained only on predicting 15-token-long fragments. Notably, our experiments are conducted on relatively short text generation (whose length statistics are shown in Table \ref{tab:datasets}), and the performance on longer text generation tasks requires further investigation.

(3) Compared with AR models, PreDAT requires more GPU memory during inference and takes more time in fine-tuning (typically 2$\sim$4 times in our experiments). This is because PreDAT's decoder has to process a much longer sequence.

\section{Conclusion}

In this paper, we propose a pre-train\-ing task to promote sentence-level consistency and bidirectional dependencies for NAR generation.
We demonstrate that combining the state-of-the-art NAR models with appropriate pre-training can lead to efficient and high-quality text generation on a wide range of tasks,
where our \methodname largely outperforms previous NAR pre-trained models in generation quality.
We further show that, compared with AR models, \methodname alleviates error accumulation and enhances relevance to inputs, but still introduces non-negligible grammatical problems, thereby providing new insights into the strengths and weaknesses of NAR generation.

\section*{Acknowledgments}

This paper was supported by the National Science Foundation for Distinguished Young Scholars (with No. 62125604) and the Guoqiang Institute of Tsinghua University, with Grant No. 2020GQG0005. 
We are grateful to the action editor, Alexander Rush, and the anonymous reviewers for their valuable suggestions and feedback.

\bibliography{tacl2021}
\bibliographystyle{acl_natbib}

\iftaclpubformat

\onecolumn

\fi

\end{document}